\documentclass[lettersize,journal]{IEEEtran}
\usepackage{amsmath,amsfonts}
\usepackage{array}
\usepackage[caption=false,font=normalsize,labelfont=sf,textfont=sf]{subfig}
\usepackage{textcomp}
\usepackage{stfloats}
\usepackage{url}

\usepackage[colorlinks, linkcolor=red,  anchorcolor=blue, citecolor=blue]{hyperref}

\usepackage{algorithm}%
\usepackage[utf8]{inputenc}
\usepackage{algpseudocode}

\usepackage{booktabs} 
\usepackage{multirow} 
\usepackage{array}
\usepackage{tabularx}
\usepackage{pifont}   
\usepackage{xcolor}

\usepackage{verbatim}
\usepackage{graphicx}
\usepackage{multirow}
\usepackage{makecell}
\usepackage{amssymb}
\usepackage{times}
\usepackage{epsfig}
\usepackage{booktabs}
\usepackage{flushend}
\usepackage{xcolor}
 
\def\BibTeX{{\rm B\kern-.05em{\sc i\kern-.025em b}\kern-.08em
    T\kern-.1667em\lower.7ex\hbox{E}\kern-.125emX}}
\usepackage{balance}

\begin{document}
\title{Contrastive Learning Guided  Latent Diffusion Model for Image-to-Image Translation}
\author{\makecell{Qi~Si, Bo~Wang, Zhao~Zhang,~\IEEEmembership{Senior Member,~IEEE}, Mingbo~Zhao, Xiaojie~Jin,\\ Mingliang~Xu and Meng~Wang,~\IEEEmembership{Fellow, IEEE}}

\thanks{
Qi Si and Meng Wang are with the School of Computer Science and Information Engineering, Hefei University of Technology, Hefei 230601, China (e-mail: shardenq@gmail.com; eric.mengwang@gmail.com).

Bo Wang is with the School of Computer Science and Information Engineering, Hefei University of Technology, Hefei 230601, China (e-mail: runbor1993@gmail.com).

Zhao Zhang is with the School of Computer Science and Information Engineering, Hefei University of Technology, Hefei, China and Yunnan Key Laboratory of Software Engineering, Yunan, China (e-mail: cszzhang@gmail.com).

Mingbo Zhao is with the School of Computer Science and Engineering, Donghua University, Shanghai, China.

Xiaojie Jin is with the ByteDance, Culver City, USA.

Mingliang Xu is with the School of Computer and Artificial Intelligence, Zhengzhou University, Zhengzhou, China.

Zhao~Zhang is the corresponding author. E-mail: cszzhang@gmail.com.
}}


\maketitle

\begin{figure*}[t]
	\makeatletter
	\makeatother
	\centering
	\includegraphics[width=\linewidth]{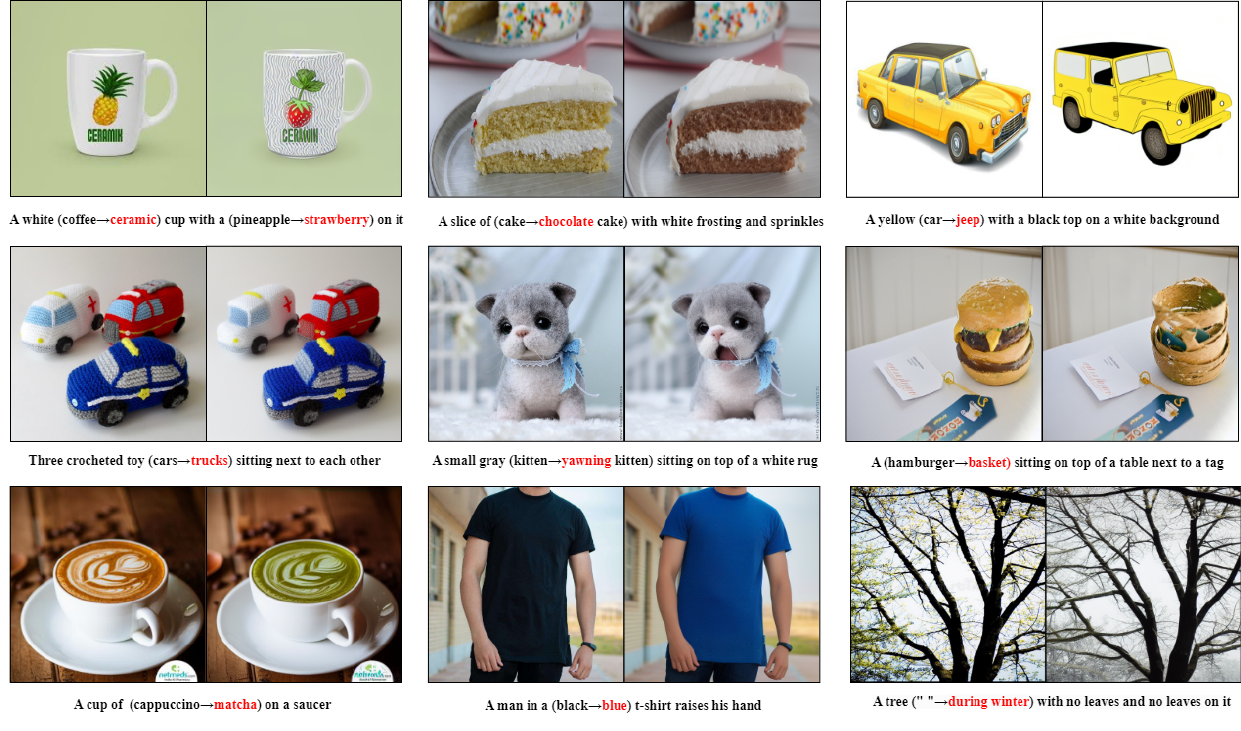}
	\caption{The effect of text-guided image translation is realized by our method. Our method successfully translates the source image, maintaining the structural elements of the source image while transforming the content according to the target text prompts.}
	\label{plot8}
\end{figure*}

\begin{abstract}
The diffusion model has demonstrated superior performance in synthesizing diverse and high-quality images for text-guided image translation. However, there remains room for improvement in both the formulation of text prompts and the preservation of reference image content. First, variations in target text prompts can significantly influence the quality of the generated images, and it is often challenging for users to craft an optimal prompt that fully captures the content of the input image. Second, while existing models can introduce desired modifications to specific regions of the reference image, they frequently induce unintended alterations in areas that should remain unchanged. To address these challenges, we propose pix2pix-zeroCon, a zero-shot diffusion-based method that eliminates the need for additional training by leveraging patch-wise contrastive loss. Specifically, we automatically determine the editing direction in the text embedding space based on the reference image and target prompts. Furthermore, to ensure precise content and structural preservation in the edited image, we introduce cross-attention guiding loss and patch-wise contrastive loss between the generated and original image embeddings within a pre-trained diffusion model. Notably, our approach requires no additional training and operates directly on a pre-trained text-to-image diffusion model. Extensive experiments demonstrate that our method surpasses existing models in image-to-image translation, achieving enhanced fidelity and controllability.	
\end{abstract}

\section{Introduction}
Diffusion modeling (DM) has made significant advances in controlled multimodal generation tasks. Notably, the latent diffusion model (LDM), an extension of the foundational diffusion model, has proven highly effective in text-to-image (T2I) tasks ~\cite{LDM1,LDM2,sd,sd1,sd2,sd3}. Building on this success, various approaches have been proposed to adapt diffusion models for image editing tasks conditioned on text. Early efforts primarily focused on incorporating source image conditions into the sampling or inversion process, or directly training models that use the source image as a conditional input ~\cite{2}. With the advent of LDM, researchers have increasingly leveraged its enhanced capabilities for T2I tasks.

In recent years, numerous diffusion model-based approaches have been proposed for image-to-image translation tasks. The majority of these methods adopt a training-free paradigm, with notable examples including plug-and-play (pnp)~\cite{pnp}, prompt-to-prompt (p2p)~\cite{P2P}, and pix2pix-zero (p2p-zero) ~\cite{p2pzero}. The pnp model achieves fine-grained control over generative structures by manipulating latent space features within the self-attention mechanism of the diffusion model. In contrast, p2p is an intuitive editing technique that facilitates the manipulation of both local and global details in images generated by text-driven diffusion models. p2p-zero introduces an automatic learning approach for the editing direction in the text embedding space, along with a cross-attention map loss to preserve the structural integrity of the original image. However, due to reconstruction errors in both the forward and reverse processes of the diffusion model, p2p-zero's ability to retain image details—such as content and structure—remains suboptimal and requires further enhancement.

To fully preserve the structure and content of the reference image in the generated output, several studies have integrated contrastive learning into the diffusion model framework. Contrastive Unpaired Translation (CUT)~\cite{19}  maximizes the mutual information between corresponding input and output patches in the latent space. Building on the success of these techniques, similar contrastive learning strategies have been adapted for pixel-domain diffusion models ~\cite{28}, employing the attention layer from ViT~\cite{DiffuseIT}  features or score networks as input for the CUT loss. Likewise, CUT loss has been applied to latent diffusion models. For instance, the Contrastive Denoising Score (CDS)~\cite{DDS} model incorporates CUT loss within the Delta Denoising Score (DDS)~\cite{CDS} framework, exploiting the rich spatial information provided by the self-attention layer in the diffusion model. However, none of these approaches can automatically generate the corresponding textual prompts for the reference image, thus significantly increasing the user’s workload.

To address the aforementioned challenges, we propose a novel training-free diffusion model method for image-to-image translation tasks, termed pix2pix-zeroCon. The main results of the proposed method are demonstrated in Fig.~\ref{plot8}. Our approach is related to, yet distinct from, the p2p-zero model. Specifically, while the cross-attention mechanism in the diffusion model is derived from the association between spatial features and text, allowing it to capture rough object-level regions in the reference image~\cite{pnp}, the p2p-zero model—relying solely on cross-attention map loss—fails to preserve the content structure of the reference image on certain datasets. To effectively resolve this issue, we incorporate CUT loss into the p2p-zero framework, ensuring that both the structure and content of the reference image are faithfully retained in the generated output. The representative results of our method and p2p-zero are displayed in Fig.~\ref{tu1}. Additionally, in comparison with state-of-the-art potential diffusion models such as CDS and DDS, our method does not require a manually specified target prompt for the reference image. Instead, we leverage pre-trained BLIP~\cite{BLIP} and CLIP~\cite{CLIP} models to automatically generate the corresponding target prompts and editing directions,  which greatly reduces the user's burden, as shown in Fig.~\ref{SIM}.

In summary, the key contributions of this paper are as follows:
\begin{itemize}
	\item \textbf{Content Structure Preservation Loss}: We observe that the cross-attention map alone in the p2p-zero model is insufficient to fully preserve the content structure of the reference image. To address this, we introduce the CUT loss, inspired by contrastive learning, which takes the latent space embeddings of both the reference and target images as input.
	
	\item \textbf{Effective Editing Direction Strategy}: To reduce the user's effort, we propose a strategy for automatically generating the target prompt as well as the editing direction. This method is applicable to a wide range of input images, and the editing directions are derived from multiple sentences, making the approach more robust than methods that rely solely on finding the direction between the original and edited words.
	
	\item \textbf{Experimental Validation}: We conduct extensive experiments to validate the effectiveness of our approach across various image translation tasks, including target change, landscape change, and face modification, among others.
\end{itemize}

\section{Related Work}
\subsection{Image-to-Image Translation}
Image-to-image (I2I) translation tasks aim to map an image from the source domain to the target domain while preserving essential features, such as the structure of the target object, in the reference image. This approach has evolved from traditional methods to modern, data-driven deep learning techniques and is widely employed to solve various vision-related problems~\cite{7,10,17}. GAN-based deep learning models are commonly used to align the output image with the target domain’s distributions~\cite{23,29,30}, though these models typically require task-specific training. Furthermore, several studies have explored the use of pre-trained GANs to perform transformations within the latent space or enable zero-shot learning with a single image pair~\cite{1,45,35,46}. With the advancement of unconditional diffusion models, these approaches have also been applied to I2I tasks, offering promising alternatives for image translation ~\cite{39,48}. In this paper, we propose a novel text-guided diffusion-based model for I2I translation, where the target domain is specified by textual prompts rather than image sets. This zero-shot method requires no training and is adaptable to a wide range of I2I tasks, such as object modification, landscape transformation, and facial expression change. The proposed approach offers significant flexibility, enabling diverse practical applications.

\subsection{Text-guided Image Manipulation} 
During image editing and translation, the primary objective is to transform the semantic content while preserving the structure and key features of the source image. To achieve this, CycleGAN~\cite{SG} first introduced the concept of cyclic consistency, which allows the output image to be translated back to the source domain, thereby maintaining the essential features of the original image. Subsequent studies    have expanded upon this concept, proposing various consistency regularization techniques to improve the effectiveness of image-to-image translation (I2I) models~\cite{CDS3,CDS2,CDS1}. On a different front, Contrastive Unpaired Translation (CUT) introduced the novel application of contrastive learning to patch-wise representations~\cite{19}, effectively preserving structural consistency between source and target images. Inspired by this, researchers have applied CUT loss to models such as StyleGAN~\cite{14} and diffusion models~\cite{28}, achieving impressive results. In this paper, we propose a novel, training-free approach that seamlessly integrates CUT loss into the p2p-zero framework, utilizing off-the-shelf latent diffusion models (LDMs). This zero-shot integration significantly enhances the quality and effectiveness of the image editing output compared to the traditional p2p-zero method. Moreover, this approach avoids the complex model training process, making it possible to improve the quality of image editing and conversion without increasing the computational burden.

\section{Preliminaries}

In the framework of diffusion models~\cite{sd,sd1}, a forward Markovian noising process is employed to progressively introduce noise to the initial input image, denoted as $\boldsymbol{x}_0$, over multiple time steps $t$. This process is mathematically expressed as
\begin{equation}
	q(\boldsymbol{x}_t|\boldsymbol{x}_{t-1}) = \mathcal{N}(\boldsymbol{x}_t; \sqrt{\bar{\alpha}_t} \boldsymbol{x}_{t-1}, (1 - \alpha_t) \mathbf{I}),
\end{equation}
\begin{equation}
	q(\boldsymbol{x}_{1:T}|\boldsymbol{x}_0) = \prod_{t=1}^T q(\boldsymbol{x}_t|\boldsymbol{x}_{t-1}),
\end{equation}
where $\alpha_t$ denotes the variance controlling the noise schedule, and $T$ refers to the total number of steps. As $T$ increases sufficiently, $\boldsymbol{x}_T$ will asymptotically approach standard Gaussian noise.

Due to the independence of transitions in a Markov process, $\boldsymbol{x}_t$ can be directly derived by introducing noise to $\boldsymbol{x}_0$ , as follows
\begin{equation}
	q(\boldsymbol{x}_t|\boldsymbol{x}_0) = \mathcal{N}(\boldsymbol{x}_t; \sqrt{\bar{\alpha}_t} \boldsymbol{x}_0, (1 - \bar{\alpha}_t) \mathbf{I}),
\end{equation}
\begin{equation}
	\boldsymbol{x}_t = \sqrt{\bar{\alpha}_t} \boldsymbol{x}_0 + \sqrt{1 - \bar{\alpha}_t} \boldsymbol{\epsilon}_t,
\end{equation}
where $\bar{\alpha}_t = \prod_{s=1}^t \alpha_s$. The conditional probability distribution, derived by Ho et al.~\cite{DDPM}, describes the reversal of the process and is given by

\begin{equation}
	\boldsymbol{p}_\theta(\boldsymbol{x}_{t-1}|\boldsymbol{x}_t) = \boldsymbol{\mathcal{N}}(\boldsymbol{\mu}_\theta(\boldsymbol{x}_t, t), \boldsymbol{\Sigma}_\theta(\boldsymbol{x}_t, t)),
\end{equation}
\begin{equation}\label{4}
	\boldsymbol{\mu}_\theta(\boldsymbol{x}_t, t) = \frac{1}{\sqrt{\bar{\alpha}_t}} (\boldsymbol{x}_t - \frac{\beta_t}{1 - \bar{\alpha}_t} \boldsymbol{\epsilon}_t),
\end{equation}
where $\beta_t = 1 - \alpha_t$. To estimate the noise, a deep neural network $\beta_t = 1 - \alpha_t$ is trained to predict $\boldsymbol{\epsilon}_t$ for a given noise level $\boldsymbol{x}_t$, utilizing the following loss function

\begin{equation}
	\mathcal{L} = \mathbb{E}_{t \sim [1, T], \boldsymbol{x}_0, \boldsymbol{\epsilon}_t} \| \boldsymbol{\epsilon}_t - \boldsymbol{\epsilon}_\theta(\boldsymbol{x}_t, t) \|^2.
\end{equation}

Combined with Eq.~(\ref{4}), we can also obtain the predicted input image by one reverse step, i.e.,

\begin{equation}
	\boldsymbol{x}_{0,t} = \frac{\boldsymbol{x}_t - \sqrt{1 - \bar{\alpha}_t} \boldsymbol{\epsilon}_\theta(\boldsymbol{x}_t, t)}{\sqrt{\bar{\alpha}_t}}.
\end{equation}

In the following discussions, we refer to $\boldsymbol{x}_{0,t}$ as the one-step reconstruction. As the denoising process advances, $\boldsymbol{x}_{0,t}$ gradually becomes a closer approximation of the input data, eventually converging to the denoising endpoint where it exactly matches $\boldsymbol{x}_0$.

\begin{figure*} 
	\centering
	\includegraphics[width=\textwidth]{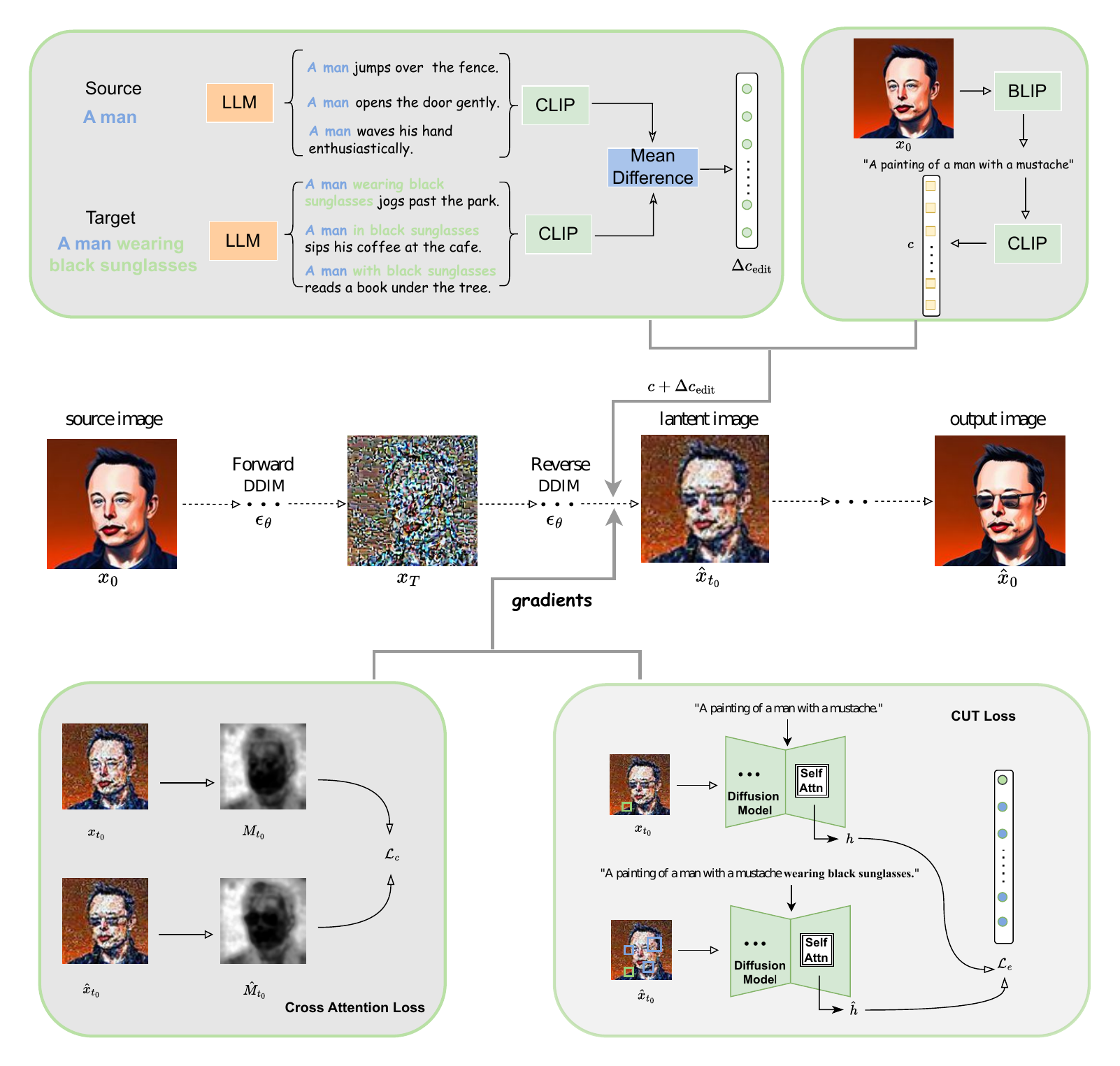}
	\caption{Pipeline of the proposed method: To guide the diffusion model, we first obtain the text embedding and editing directions using the pre-trained BLIP and CLIP models, respectively. We then employ the cross-attention loss and CUT loss to steer the denoising process of the diffusion model.}
	\label{pipe}
\end{figure*}

\section{Method}
There are two parts to our method, refer to Fig.~\ref{pipe} for the presentation of the specific pipeline. First, the input image $\boldsymbol{x}_0$ undergoes pre-processing, which includes the diffusion forward process, encoding of the input image, and extraction of the editing direction, as described in Section~\ref{3.1}. Additionally, the reverse denoising process of the diffusion model is guided by the design of effective loss functions, with further details provided in Section~\ref{3.2}. 
\subsection{Input Pre-processing Process}\label{3.1}
\begin{figure} 
	\centering
	\includegraphics[width=\columnwidth]{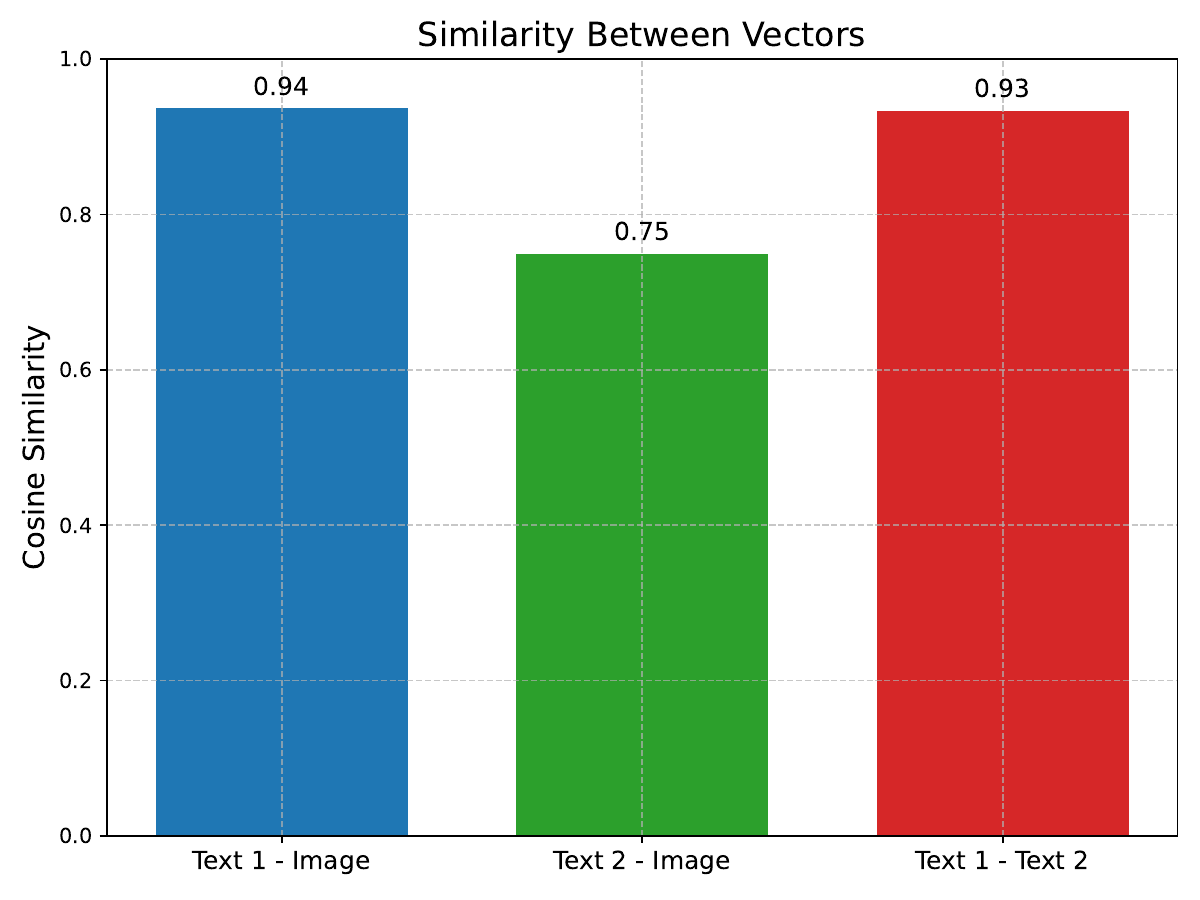}
	\caption{Comparison of the similarity between the encoded text vectors and image vectors obtained before and after adding the editing direction $\Delta \boldsymbol{c}$, where text and image correspond to the examples given in Fig.~\ref{pipe}. The ``Text 2" is obtained by encoding the target prompt ``A painting of a man with a mustache wearing black sunglasses". The ``Text 1" is obtained by encoding the source prompt ``A painting of a man with a mustache" and adding the edit direction $\Delta \boldsymbol{c}$. The ``Image" is obtained by encoding the source image latent $\boldsymbol{x}_0$.}
	\label{SIM}
\end{figure}

\noindent \textbf{Deterministic Inversion}. The inversion process aims to identify a noise map $\boldsymbol{x}_T$, where $T$ represents the total number of noise steps. This noise map $\boldsymbol{x}_T$ is capable of reconstructing the input latent $\boldsymbol{x}_0$ during the reverse process. In DDPM~\cite{DDPM}, this corresponds to a forward additive noise process, followed by denoising in the reverse process. However, both the forward and reverse processes in DDPM are inherently stochastic, which limits the quality of the reconstruction. To mitigate this issue, we adopt the deterministic reverse process from DDIM~\cite{DDIM}, the mathematical formulation of which is provided below
\begin{equation}
	\boldsymbol{x}_{t+1} = \sqrt{\bar{\alpha}_{t+1}} \boldsymbol{f}_{\theta}(\boldsymbol{x}_t,t,c) + \sqrt{1 - \bar{\alpha}_t} \boldsymbol{\epsilon}_{\theta}(\boldsymbol{x}_t,t,c),
\end{equation}
where $\boldsymbol{x}_t$ represents the noisy latent code at time step $t$, $\boldsymbol{\epsilon}_\theta(\boldsymbol{x}_t, t, \boldsymbol{c})$ is a U-Net-based denoiser that predicts the noise in $\boldsymbol{x}_t$ based on the time step $t$ and the encoded text feature $\boldsymbol{c}$, $\bar{\alpha}_{t+1}$ is the noise scaling factor defined in DDIM, and $\boldsymbol{f}_\theta(\boldsymbol{x}_t, t, \boldsymbol{c})$ denotes the final prediction of the latent code, which can be expressed as follows
\begin{equation}
	\boldsymbol{f}_\theta(\boldsymbol{x}_t, t, c) = \frac{\boldsymbol{x}_t - \sqrt{1 - \bar{\alpha}_t} \boldsymbol{\epsilon}_\theta(\boldsymbol{x}_t, t, c)}{\sqrt{\bar{\alpha}_t}}.
\end{equation}
\noindent \textbf{Editing Direction}. The recently developed multi-modal large models enable users to obtain corresponding textual descriptions by inputting an image, offering significant convenience. In contrast, we aim to provide users with an interface that only requires them to specify the desired text change from the source domain to the target domain. For instance, the text modification might be: 'A man' → 'A man with glasses.' To determine the direction of editing in the latent space, we first encode the input image $\boldsymbol{x}_0$ using the pre-trained BLIP model~\cite{BLIP}, as represented by the following mathematical expression
\begin{equation}
P=\text{BLIP}(\boldsymbol{X}_0),
\end{equation}
where $P$ is the source prompt, and the target prompt $\hat{P}$ can be derived based on the user-supplied editing text. We then utilize an out-of-the-box sentence generator, such as GPT-3, to generate a large number of diverse sentences based on the source and target prompts~\cite{GPT3}, denoted as $\{P_1, P_2, \ldots, P_n\}$ and $\{\hat{P}_1, \hat{P}_2, \ldots, \hat{P}_n\}$. Next, we employ the pre-trained CLIP model to compute the average difference between the source and target sentences~\cite{CLIP}, as expressed in the following mathematical formulation
\begin{equation}
	\Delta \boldsymbol{c} = \frac{\sum_{i=1}^{n} \text{CLIP}(P_i) - \sum_{i=1}^{n} \text{CLIP}(\hat{P}_i)}{n},
\end{equation}
where $n$ represents the total number of source or target sentences generated. After obtaining the edit direction $\Delta \boldsymbol{c}$, we use $\boldsymbol{c} + \Delta \boldsymbol{c}$ as the target text embedding $\hat{\boldsymbol{c}}$, which is required in the final reverse denoising process. It is important to note that the reverse denoising process and editing operations described in this paper are performed in the latent space. For example, the input image $\boldsymbol{X}_0 \in \mathbb{R}^{X \times X \times 3}$ is first encoded into an image latent code $\boldsymbol{x}_0 \in \mathbb{R}^{X \times X \times 4}$ before being processed by the diffusion model’s denoising mechanism. In the experiments, $X = 512$ represents the image size.  During our experiments, we also observed that using multiple sentences to determine the text direction yields more robust results than relying on a single sentence. For concrete examples, refer to Fig.~\ref{SIM}, by adding the edit direction $\Delta \boldsymbol{c}$ obtained from the pre-trained language model to the source prompt embedding, it is advantageous to obtain a target prompt embedding that is more similar to the source image latent. 

\subsection{Attention Guidance}\label{3.2}
\noindent \textbf{Attention Mechanisms}. It is well known that the diffusion model consists of a denoising U-Net and a series of fundamental modules, including the residual model, the self-attention mechanism, and the cross-attention mechanism. Many recent studies have focused on improving the generative performance of diffusion models through innovations in the attention mechanism~\cite{p2pzero, masatrol, freecustom}. Prior to inference with the diffusion model, both the source image and the target prompt are first encoded and mapped into the latent space. The resulting latent codes are then passed through the corresponding modules. We employ the open-source Latent Diffusion Model (LDM)~\cite{LDM1}, whose attention mechanism module can be unified by the following mathematical expression
\begin{equation}\label{13}
	\begin{split}
		\text{Attention}(\boldsymbol{Q}, \boldsymbol{K}, \boldsymbol{V}) &= \boldsymbol{M} \cdot \boldsymbol{V}, \\
		\text{where} \quad \boldsymbol{M} &= \text{Softmax}\left( \frac{\boldsymbol{Q}\boldsymbol{K}^{\top}}{\sqrt{d}} \right),
	\end{split}
\end{equation}
where $\boldsymbol{M}$ denotes the attention map weight matrix, $\boldsymbol{Q}$ represents the query features derived from the spatial features, and $\boldsymbol{K}$ and $\boldsymbol{V}$ correspond to the key and value features, respectively. The key $\boldsymbol{K}$ and value $\boldsymbol{V}$ are obtained by mapping either the spatial features (in the self-attention layer) or the text features (in the cross-attention layer).

\noindent \textbf{Cross Attention Loss}.
Notably, the cross-attention mechanism module in the diffusion model contains rich image structure information~\cite{P2P}, which inspires us to leverage this advantage. In Eq.~(\ref{13}), $M_{i,j}$ denotes the contribution of the $j$-th text token to the $i$-th spatial location when it represents the cross-attention mechanism, and it is clear that for different denoising steps we can obtain different cross-attention map matrices $\boldsymbol{M}_t$. In the $t_0$ denoising step, we first obtain the edit direction $\hat{\boldsymbol{c}}$~(Section~\ref{3.1}) and then use this direction to guide the denoising process. However, relying solely on this edit direction is not sufficient to achieve the desired result, such as preserving the structure of the input image. Inspired by p2p-zero~\cite{p2pzero}, we introduce the cross-attention map loss to guide the denoising process and better preserve the reference image's structure.

To define the cross-attention map loss, we proceed in two steps. First, during the $t_0$ denoising step, we use the source text embedding $\boldsymbol{c}$ to obtain the cross-attention map $\boldsymbol{M}_{t_0}$ from the cross-attention module. Next, to preserve the structure of the source image, we use the obtained edit direction $\hat{\boldsymbol{c}}$ during the $t_0$ denoising step to generate a new cross-attention map $\hat{\boldsymbol{M}}_{t_0}$. To align with the reference cross-attention map $\boldsymbol{M}_{t_0}$, the cross-attention map loss is formulated as follows 
\begin{equation}\label{jiaocha}
\mathcal{L}_{\text{c}} = \| \hat{\boldsymbol{M}}_{t_0} - \boldsymbol{M}_{t_0}\|_2.
\end{equation}

\noindent \textbf{CUT Loss}.
Due to the presence of reconstruction errors in the diffusion model, relying solely on the edit direction $\Delta \boldsymbol{c}$ and the cross-attention map loss, as described above, is not always sufficient to fully preserve the content of the reference image across certain datasets. To address this, we introduce the CUT loss. It has been shown that maximizing the mutual information between input and output image patches using CUT loss can effectively preserve the content structure of the input image~\cite{28}. Specifically, previous approaches employing CUT loss typically require an additional encoder to be trained in order to capture the spatial information of the input image effectively. In contrast, a recent study by Baranchuk et al.~\cite{2} demonstrated that the U-Net noise predictor in the diffusion model contains rich spatial information, which aligns perfectly with the requirements of CUT loss. Therefore, in this work, we introduce CUT loss to further preserve the content structure of the reference image without the need for additional training.

Specifically, we first describe the relationship between the self-attention layer in the denoising U-Net $\boldsymbol{\epsilon}_\theta$ and the CUT loss. At each inverse timestep $t_0$, both the source image latent code $\boldsymbol{x}_{t_0}$ and the reverse-sampled denoised image latent code $\hat{\boldsymbol{x}}_{t_0}$ are passed to the noise estimator $\boldsymbol{\epsilon}_\theta$. The encoder component of the estimator extracts the feature maps $\boldsymbol{h}_l$ and $\hat{\boldsymbol{h}}_l$ for $\boldsymbol{x}_{t_0}$ and $\hat{\boldsymbol{x}}_{t_0}$, respectively. $\hat{\boldsymbol{h}}_l$ and $\boldsymbol{h}_l$ represent the intermediate features passing through the $l$-th layer of the residual block and the self-attention block, respectively. Initially, a ``query" patch $\boldsymbol{h}_l^s$  is randomly selected from the feature map $\hat{\boldsymbol{h}}_l$. We define $s \in \{1, \dots, S_l\}$, where $S_l$ is the number of query patches. For each query, the patch at the corresponding position in $\boldsymbol{h}_l$ is also selected. Patches at the same position in the feature map $\hat{\boldsymbol{h}}_l$ are referred to as ``positive" patches, while patches at non-identical positions are referred to as ``negative" patches. The objective of the CUT loss is to maximize the mutual information between ``positive" patches while minimizing the mutual information between ``negative" patches. This process is expressed by the CUT loss as follows
\begin{equation}
	\mathcal{L}_{e}(\boldsymbol{x}_{t_0}, \hat{\boldsymbol{x}}_{t_0}) = \mathbb{E}_{\boldsymbol{h}} \left[ \sum_l \sum_s \ell(\boldsymbol{h}_l^s, \hat{\boldsymbol{h}}_l^s, \hat{\boldsymbol{h}}_l^{S\setminus s}) \right]
\end{equation}
where $\boldsymbol{h}_l^s$ is a ``query" patch, with positive patches denoted as  $\hat{\boldsymbol{h}}_l^s$ and ``negative" patches denoted as $\hat{\boldsymbol{h}}_l^{S \setminus s}$. $(\cdot)$ represents the cross-entropy loss, its specific mathematical expression is as follows 
\begin{equation}
	\ell(\boldsymbol{h}, \boldsymbol{h}^+, \boldsymbol{h}^-) = -\log \left[
	\frac{\exp(\boldsymbol{h} \cdot \boldsymbol{h}^+ / \tau)}
	{\exp(\boldsymbol{h} \cdot \boldsymbol{h}^+ / \tau) + 
		\sum\limits_j \exp(\boldsymbol{h} \cdot \boldsymbol{h}_j^- / \tau)}
	\right], 
\end{equation}
where $\tau$ is a temperature parameter.

The overall guidance loss function is defined as
\begin{equation}
	\mathcal{L}_{t} = \lambda_c\mathcal{L}_c+\lambda_e\mathcal{L}_e,
\end{equation}
where $\lambda_c$ and $\lambda_e$ are attention guidance weights. 

Then, we compute the gradient of latent map $\boldsymbol{x}_{t_0}$ and update the predicted latent map $\boldsymbol{x}_{t_0}$ as
\begin{equation}
	\boldsymbol{x}_{t_0} \leftarrow \boldsymbol{x}_{t_0}-\lambda\nabla_{\boldsymbol{x}_{t_0}}\mathcal{L}_{t},
\end{equation}
where $\lambda$ is the learning rate parameter.

\begin{algorithm}[!htb]
	\caption{pix2pix-zeroCon}
	\label{suanfa1}
	\begin{algorithmic}[1]
		\Require
		\Statex The source text embedding  $\boldsymbol{c}$, the edit direction $\Delta \boldsymbol{c}$, a source image lantent code $\boldsymbol{x}_0$, the initial latent noise $\boldsymbol{x}_T$ corresponding to $\boldsymbol{x}_0$, the denoising step $T$, the attention guidance weights $\lambda_c$ and $\lambda_e$, the learning rate $\lambda$.
		\Ensure
		\Statex The edited latent map $\hat{\boldsymbol{x}}_0$.
		\Statex $\triangleright$ Compute source cross-attention maps
		\For {$t_{0}=T,T-1,...,1$}
		\State $\boldsymbol{M}_{t_{0}}\gets \boldsymbol{\epsilon}_{\theta}(\boldsymbol{x}_{t_{0}},t_{0},\boldsymbol{c})$ 
		\EndFor    
		
		\Statex $\triangleright$ Edit with attention guidance 
		\State $\hat{\boldsymbol{c}}=\boldsymbol{c}+\Delta \boldsymbol{c}$
		\For {$t_{0}=T,T-1,...,1$}
		\State $\hat{\boldsymbol{M}}_{t_{0}}\gets \boldsymbol{\epsilon}_{\theta}(\hat{\boldsymbol{x}}_{t_{0}},t_{0},\hat{\boldsymbol{c}})$ 
		\State $\mathcal{L}_{t} \gets \lambda_c\mathcal{L}_c+\lambda_e\mathcal{L}_e$
		\State $\Delta x_{t_0} = \nabla_{x_{t_0}} \mathcal{L}_{t}$
		\State $\hat{\boldsymbol{\epsilon}} \leftarrow \epsilon_\theta(\boldsymbol{x}_{t_0} - \lambda \Delta \boldsymbol{x}_{t_0}, t_0,\hat{\boldsymbol{c}})$
		\State $\boldsymbol{x}_{t_0-1
		} = \text{UPDATE}(\boldsymbol{x}_{t_0}, \hat{\boldsymbol{\epsilon}}, t_0)$
		\EndFor  
		
		\Statex $\triangleright$ Update current state $\boldsymbol{x}_{t_0}$ with noise prediction $\hat{\boldsymbol{\epsilon}}$
		\Function {Update}{$\boldsymbol{x}_{t_0}, \hat{\boldsymbol{\epsilon}}, t_0$}
		\State $\boldsymbol{x}_{t_0-1} = \sqrt{\alpha_{t_0-1}} \frac{\boldsymbol{x}_{t_0} - \sqrt{1-\alpha_{t_0}}\hat{\boldsymbol{\epsilon}}}{\sqrt{\alpha_{t_0}}}  + \sqrt{1-\alpha_{t_0-1}} \hat{\boldsymbol{\epsilon}}$
		\State \Return $\boldsymbol{x}_{t_0-1}$
		\EndFunction
		
	\end{algorithmic}
\end{algorithm}

\section{Experiments}
\subsection{Experimental setting}
\noindent \textbf{Implementation.}
For the implementation of experimental details, we refer to the official code of p2p-zero, which uses the pre-training weights of Stable Diffusion v1.5\footnote{\url{https://github.com/CompVis/stable-diffusion}}.  All experiments are conducted on a single NVIDIA 4090 GPU.

\noindent \textbf{Baseline methods and datasets.} Both the proposed method and the comparison methods can be applied to image-to-image translation tasks on real images, enabling the editing and control of detailed structures in these images. The benchmark methods used in the experiments include p2p-zero~\cite{p2pzero}, pnp~\cite{pnp}, DDS~\cite{DDS}, CDS~\cite{CDS}, and DDIM~\cite{DDIM}. Additionally, all datasets employed are publicly available benchmark datasets, including ImageNet-R~\cite{Image-R}, LAION-5B~\cite{LAION-5B}, and CelebAMask-HQ-512\footnote{\url{https://www.kaggle.com/datasets/vincenttamml/celebamaskhq512}}. These three datasets encompass a diverse range of image types, covering various subject domains such as facial images, real object images, sketch images, and paintings.

\subsection{Experimental Results}
\begin{figure*} 
	\centering
	\includegraphics[width=1\textwidth, height=0.95\textheight]{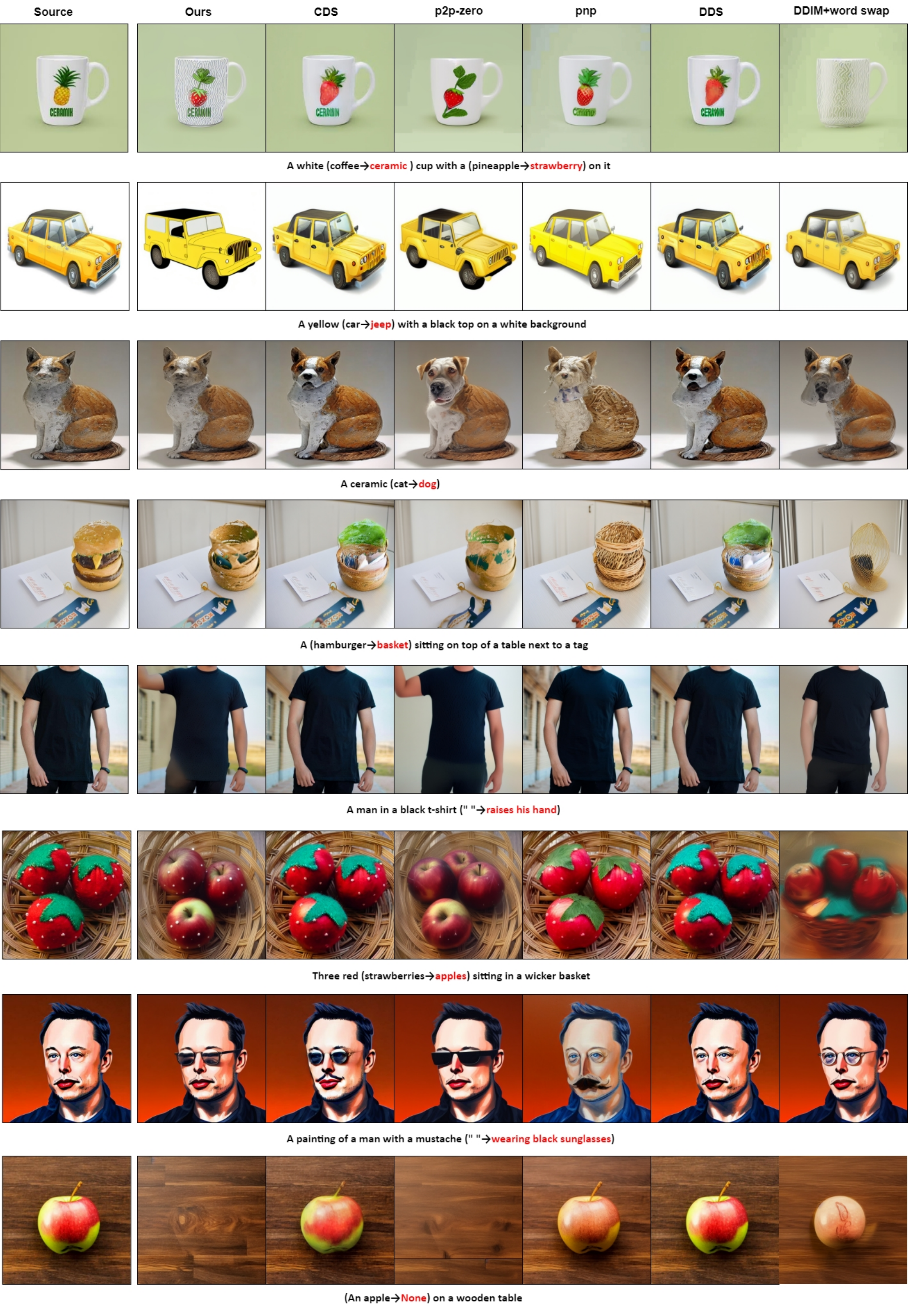}
	\caption{Comparison of qualitative experimental results of each method on real images.}
	\label{tu1}
\end{figure*}

\noindent \textbf{Qualitative Results.} 
For qualitative comparison, we present the editing results in Fig.~\ref{tu1}. The target prompt corresponding to each image in the experiment is obtained from the pre-trained BLIP model ~\cite{BLIP}.
The qualitative results shown in Fig.~\ref{tu1} reveal that the DDIM with the word swap method struggles with editing, making it difficult to preserve both the structure and content of the reference image. In contrast, the advanced methods pnp and p2p-zero, which leverage attention-based innovations, produce ideal editing results but still exhibit some limitations in preserving the background of the reference image. Although the recently developed methods CDS and DDS have advantages in background preservation, their editing results are less satisfactory. In contrast, our method outperforms all others in the qualitative results, effectively preserving the content and structure of the reference image while achieving highly desirable editing outcomes.

\begin{figure}[ht]
	\centering
	\begin{minipage}[b]{0.47\columnwidth}
		\centering
		\includegraphics[width=\textwidth]{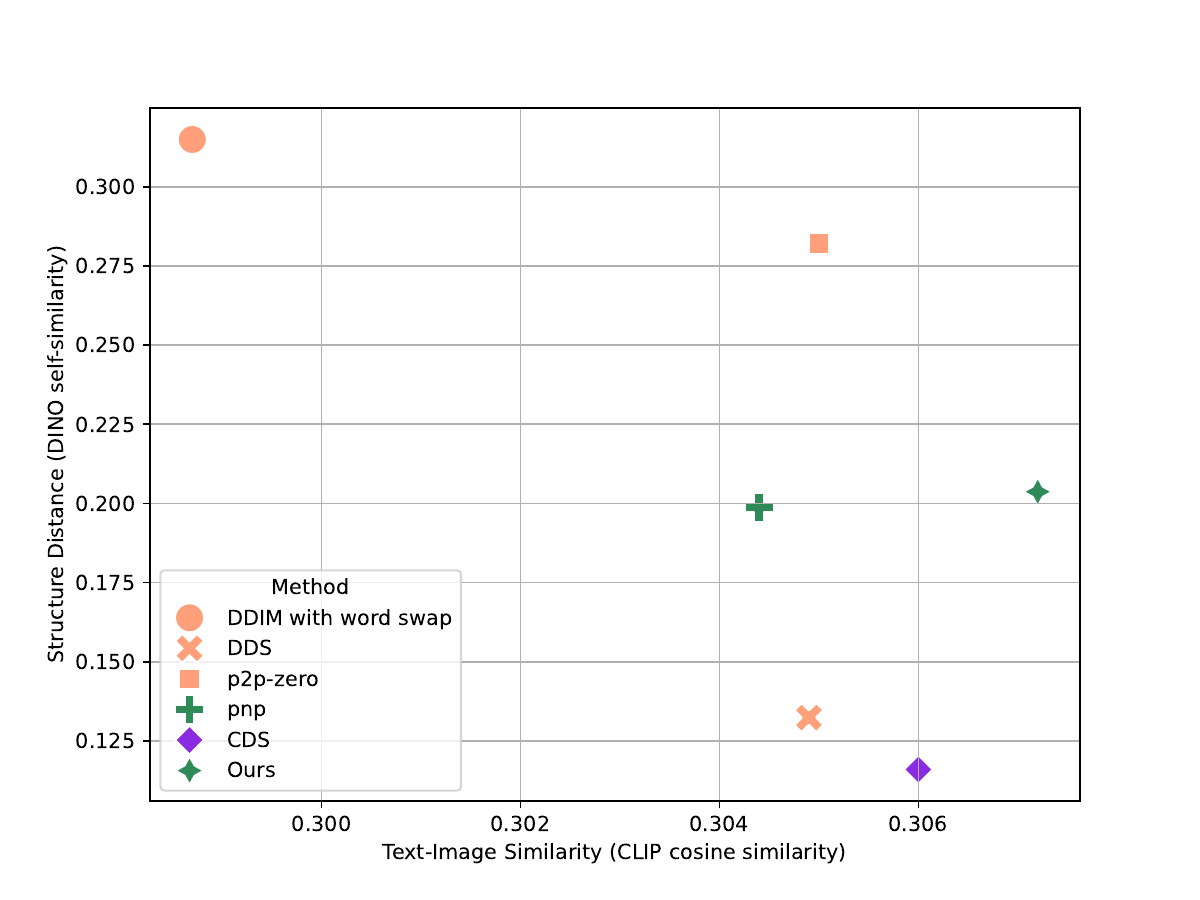}
	\end{minipage}
	\hfill
	\begin{minipage}[b]{0.47\columnwidth}
		\centering
		\includegraphics[width=\textwidth]{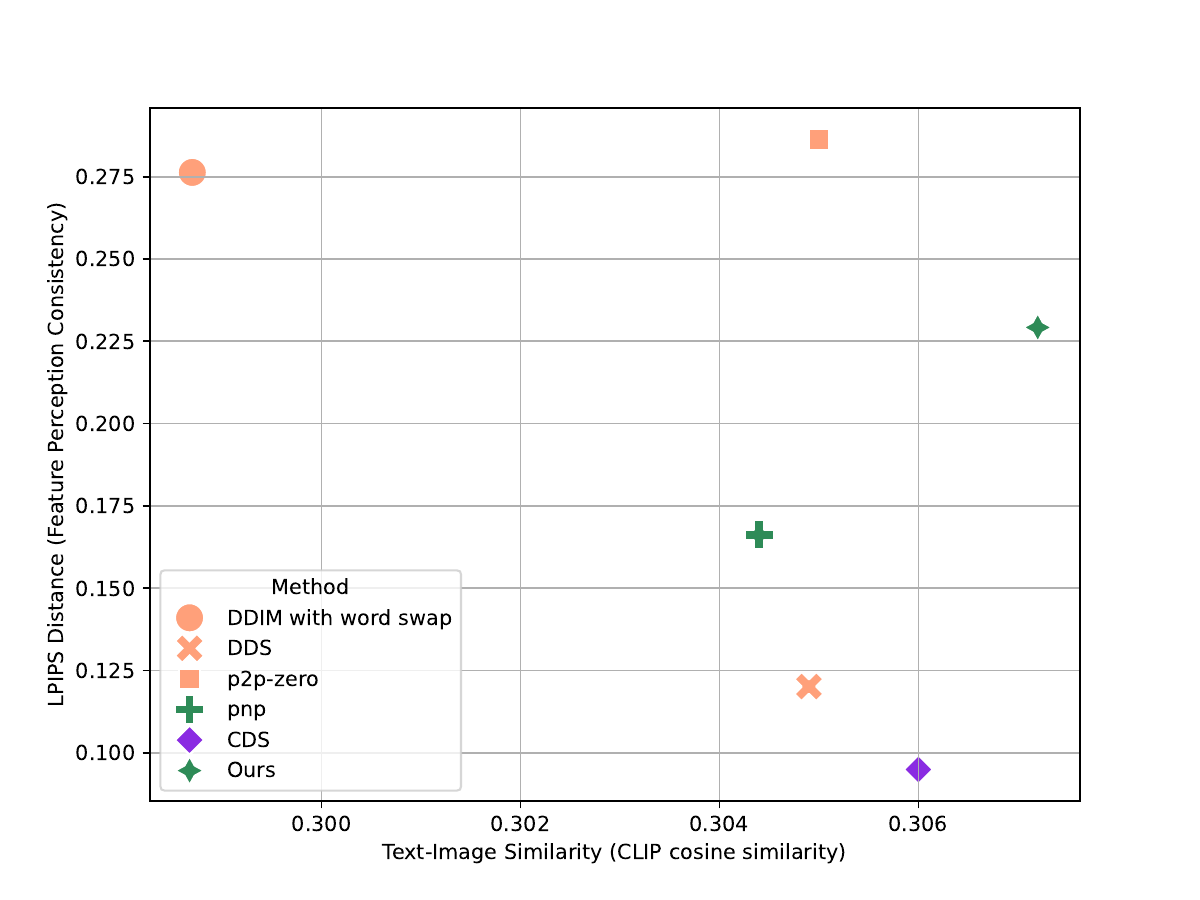}
	\end{minipage}
    \caption{Analysis of the average quantitative results corresponding to the qualitative results of each method in Fig.~\ref{tu1}.}
    \label{zhibiao}
\end{figure}

\begin{figure}
	\centering
	\includegraphics[width=\columnwidth]{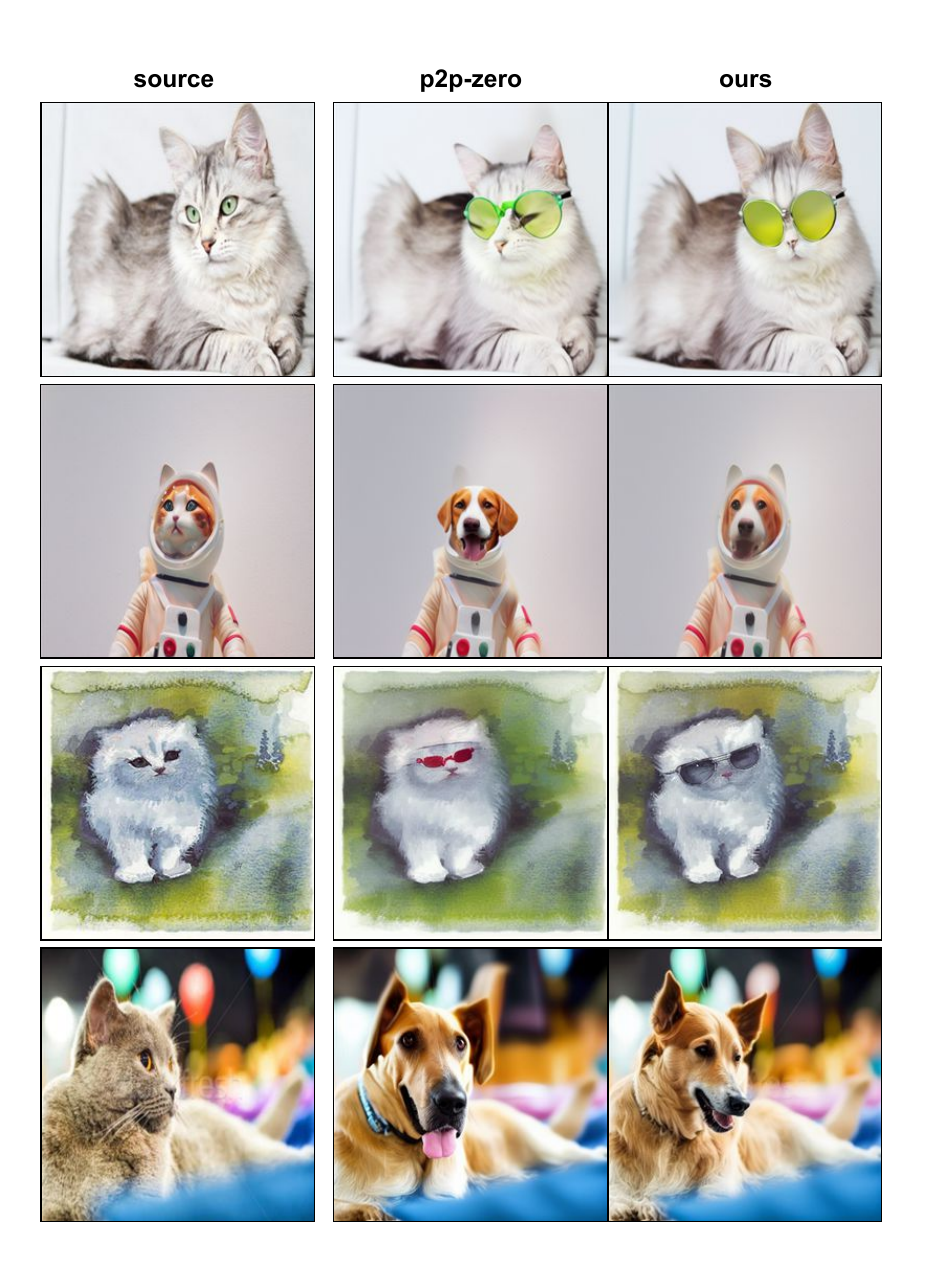}
	\caption{Representative experimental results of the proposed method with the p2p-zero method on quantitative experiments.}
	\label{tu2}
\end{figure}

\noindent \textbf{Quantitative Results.} 
In order to evaluate the generative performance of the proposed method from the quantitative view, we performed two tasks: (1) cat→dog, and (2) cat→cat with glasses. We have selected some representative examples from the above two tasks for demonstration purposes, details of which can be seen in Fig.~\ref{tu2}. Note that the DDIM method refers to the editing direction provided by word swap only. The used data sets about cats were all collected from the LAION-5B dataset~\cite{LAION-5B} and the Imagenet-R dataset~\cite{Image-R}. To assess the editing results, we first employed the CLIP model~\cite{CLIP} to measure how well the generated images align with the target prompts. Next, we used the Background Local Perceptual Image Patch Similarity (BG-LPIPS) metric to evaluate the consistency of the editing results with the content structure of the reference images~\cite{p2pzero}. Table~\ref{tab1} presents the quantitative experimental results for the proposed method and the comparison methods. The results clearly demonstrate that the proposed method outperforms the comparison methods to some extent, further highlighting its advantages in preserving the structure of the reference image and achieving superior editing performance. In addition, we also quantitatively analyzed the results of each method obtained in Fig.~\ref{zhibiao}, where we used the metrics CLIP-Score, LPIPS Distance~\cite{BGLIPIS}, and DINO-ViT Structural Distance (Dist)~\cite{Dist}. Since the last set of experiments in Fig.~\ref{tu1} is about the target removal task, we removed that set of experiments and averaged the rest of the results to obtain the final results. From the results in Fig.~\ref{zhibiao}, we can see that although the LPIPS and Dist metrics of our method are not optimal, the CLIP-Score is optimal, which is mainly due to the fact that methods such as CDS and DDS do not change the original image on some tasks (e.g., 'hand'→'raise hand' and 'man'→'man wearing the glasses'). In summary, our method achieves a good balance in terms of text-image consistency as well as structure preservation ability.

\begin{table}[ht]
	\centering
	\scriptsize
	\caption{Comparison of the quantitative results of each method on the benchmark datasets. Methods with optimal results are highlighted in black and those with sub-optimal results are underlined. }
	\label{tab1}
	\begin{tabular}{@{}lcccc@{}}
		\toprule
		\multirow{2}{*}{\textbf{Method}} & \multicolumn{2}{c}{\textbf{Cat → Dog}} & \multicolumn{2}{c}{\textbf{Cat → Cat w/ glasses}} \\
		\cmidrule(lr){2-3} \cmidrule(lr){4-5} 
		& \textbf{CLIP-Score($\uparrow$)} & \textbf{BG-LPIPS($\downarrow$)} & \textbf{CLIP-Score($\uparrow$)} & \textbf{BG-LPIPS($\downarrow$)}  \\
		\midrule
		DDIM& 0.2740 & 0.0660 & 0.3146 & 0.0910\\
		pnp & \textbf{0.3001} & 0.0649 & 0.3170& 0.0709 \\
		p2p-zero & 0.2496 & 0.0944 & \underline{0.3239} & 0.1325 \\
		DDS & 0.2641 & 0.0454 & 0.3189 & \underline{0.0582} \\
		CDS & 0.2632 & \underline{0.0294} & 0.3166 & \textbf{0.0396} \\
		Ours & \underline{0.2747} & \textbf{0.0265} & \textbf{0.3419} & 0.0744 \\
		\bottomrule
	\end{tabular}
\end{table}
	
\noindent \textbf{Comparison with p2p-zero.} To further highlight the editing performance of our method in comparison with the state-of-the-art p2p-zero method, we selected representative results from the quantitative experiments, as shown in Fig.~\ref{tu2}. The target prompts for both our method and the p2p-zero model are generated using the pre-trained BLIP model, ensuring consistency across both methods. For instance, the target prompt could be: 'A painting of a cat' → 'A painting of a cat wearing glasses.' Compared to the p2p-zero model, our method not only achieves the desired editing results but also preserves the content structure of the reference image in a more comprehensive manner. These results further validate the advantage of our method, particularly in utilizing the CUT loss to better preserve the structural integrity of the reference image.

\begin{figure} 
	\centering
	\includegraphics[width=\columnwidth]{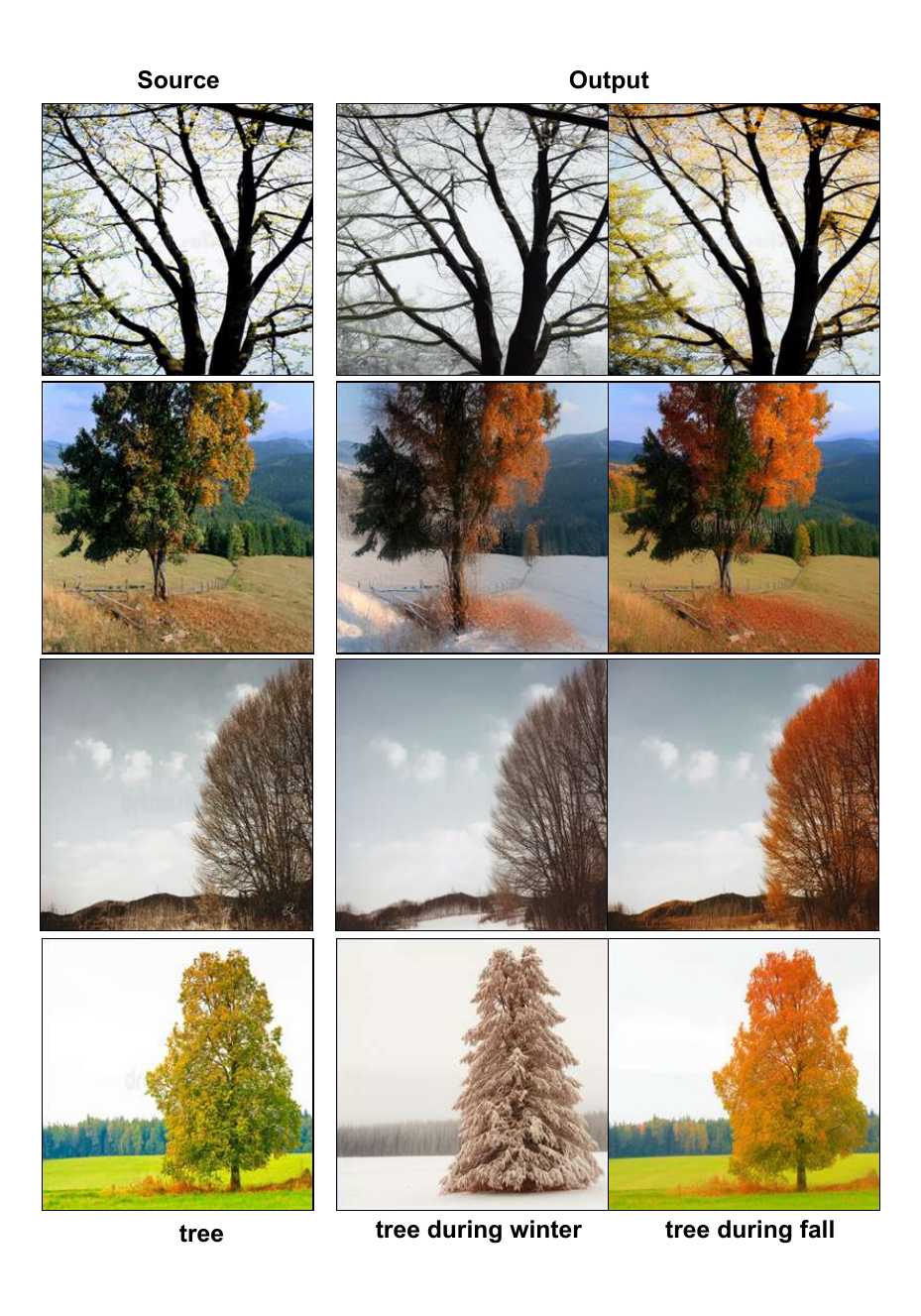}
	\caption{Representative experimental results of the proposed method on the transformation of real image scenes.}
	\label{tu4}
\end{figure}

\noindent \textbf{Scene Variation Editing.} To further demonstrate the practical application of our method, this part of the experiment showcases the results of applying our method to real-world image scenes, such as: spring-to-winter, spring-to-autumn, and so on. The results are presented in Figure~\ref{tu4}. The reference image prompt is also generated using the pre-trained BLIP model. To obtain the edited prompt, we simply add terms like 'during fall' or 'during winter' to the reference prompt (e.g., 'tree'). For instance, 'A tree with no leaves' → 'A tree during winter with no leaves.' From the results in Fig.~\ref{tu4}, it is evident that our model successfully achieves the desired editing results while preserving the overall structural layout of the reference image. This demonstrates the practicality of our model in the task of scene transformation.

\begin{figure} 
	\centering
	\includegraphics[width=\columnwidth]{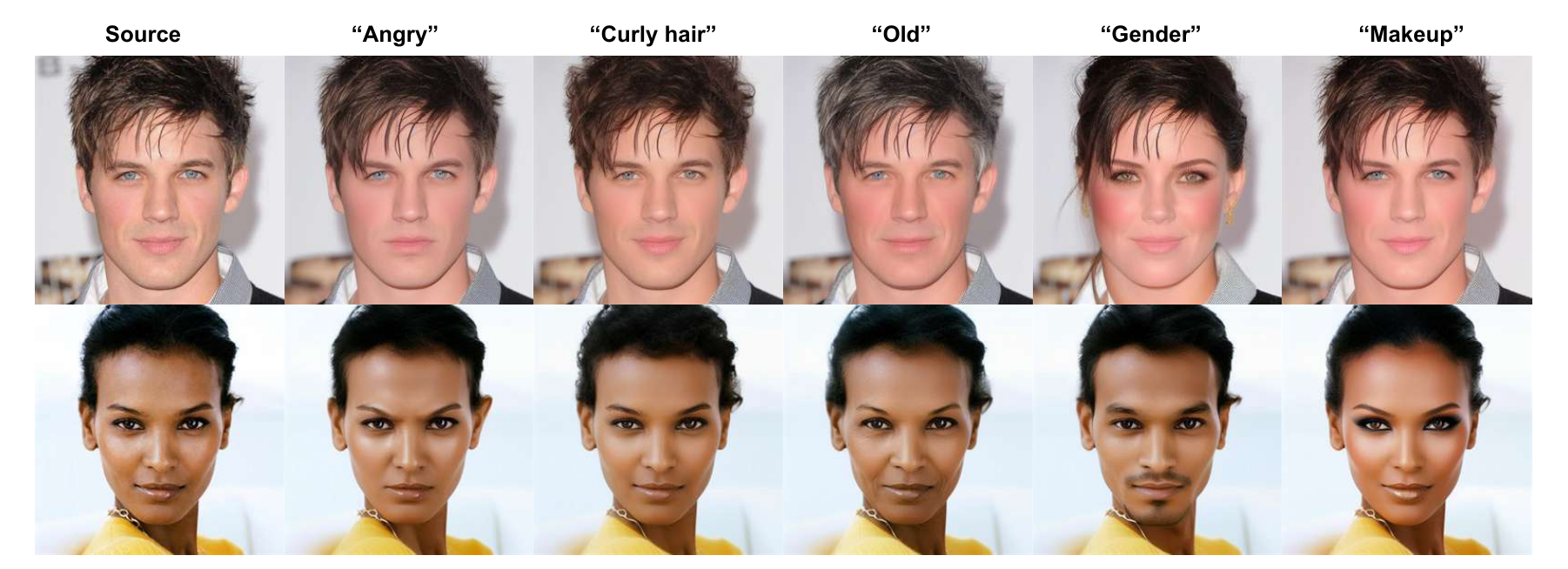}
	\caption{Representative experimental results of the proposed method on the face transformation task.}
	\label{tu6}
\end{figure}

\noindent \textbf{Face Variation Editing.}
In addition to demonstrating the advantages of our model on the previously discussed tasks, we evaluate its editing performance on the face transformation task in this part of the experiment, as shown in Fig.~\ref{tu6}. We show here six aspects of changing the face in the reference image: angry, curly, old, gender, make-up and so on. The datasets used in the experiments are all from the CelebAMask-HQ-512 dataset. Also, the target prompt in the experiment is obtained from the pre-trained BLIP model. In order to get the target prompt, we add the change text in the reference prompt, such as: angry, old, and so on. As can be seen from the results in Fig.~\ref{tu6}, our model successfully performs the editing and achieves the expected results. Thus, this part of the experiment further validates the advantages of our model in face transformation tasks.

\begin{figure} 
	\centering
	\includegraphics[width=\columnwidth]{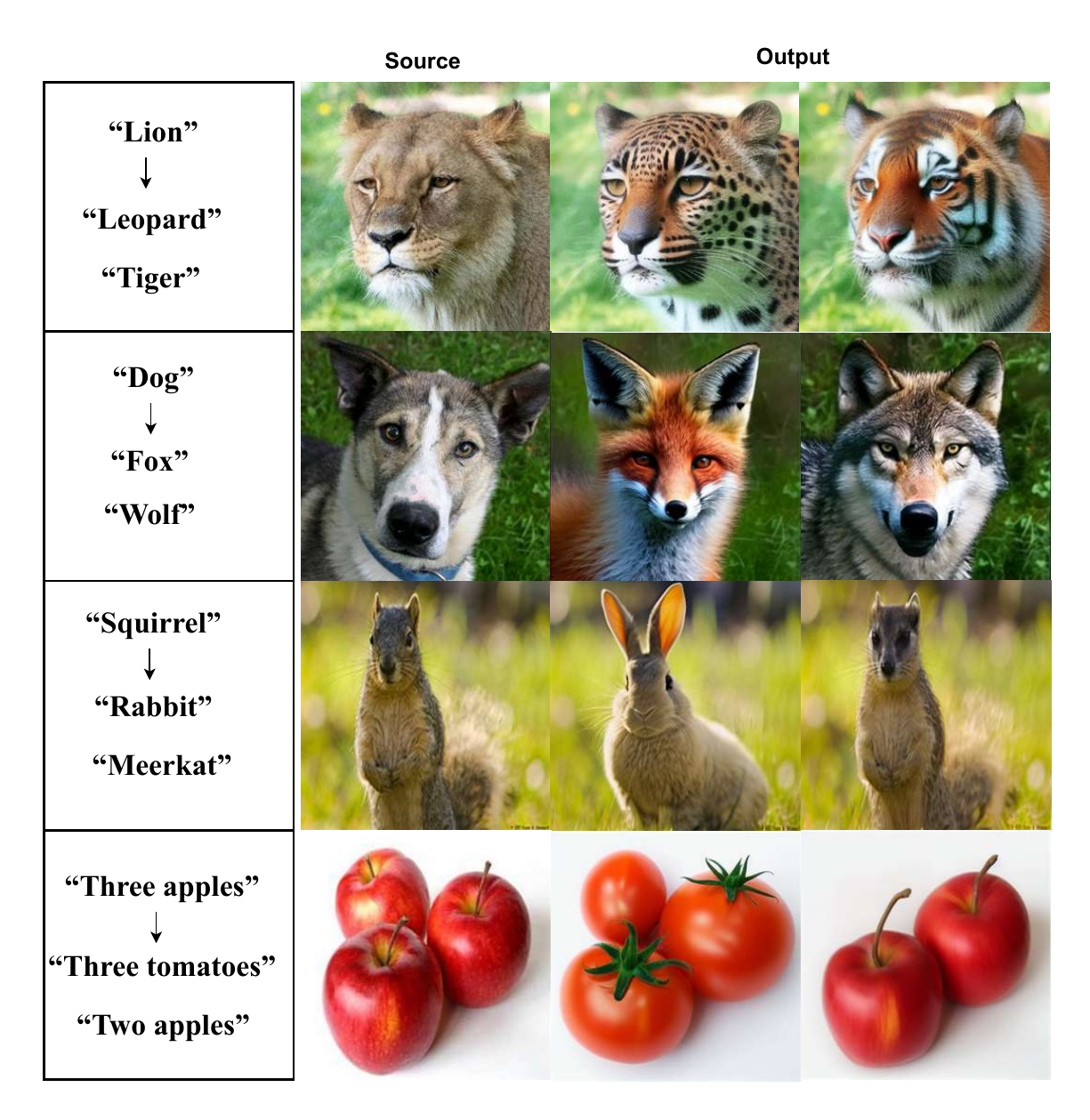}
	\caption{Representative experimental results on the proposed method image translation task.}
	\label{tu3}
\end{figure}

\noindent \textbf{Image Translation Tasks.} In this part of the experiment, we further examine the editing performance of our method on image translation tasks, selecting four representative examples for demonstration. The detailed results are presented in Fig.~\ref{tu3}. As with the previous experiments, the target prompt is initially generated by the pre-trained BLIP model and then modified according to the textual changes shown in the figures. Remarkably, our method successfully removes the target from the reference image, further validating the effectiveness of the editing directions we provide. However, a limitation is that the content structure consistent with the reference image is not fully preserved, highlighting an area for future research.

\subsection{Ablation Experiments}
\begin{figure} 
	\centering
	\includegraphics[width=\columnwidth]{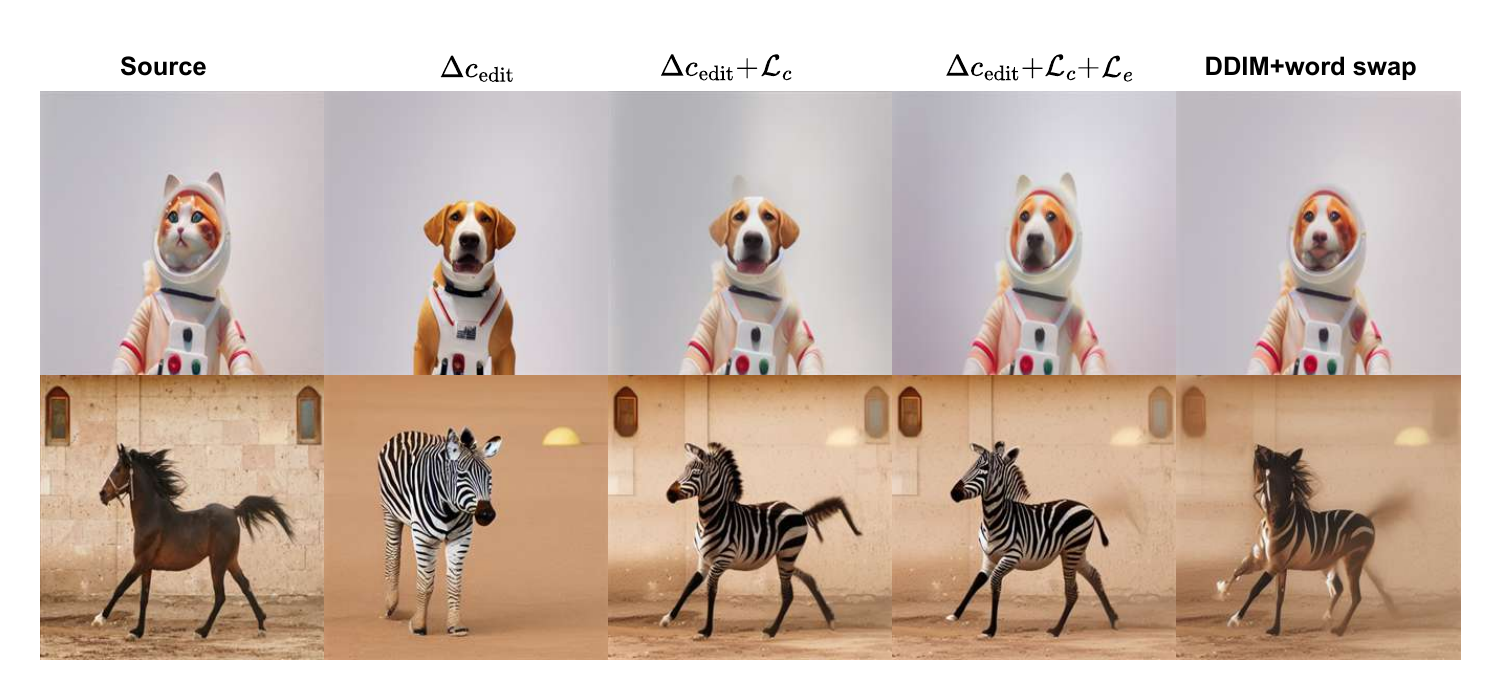}
	\caption{Results of ablation experiments of our method on two representative examples.}
	\label{tu5}
\end{figure}

\noindent \textbf{Impact of losses.} To further validate the effectiveness of our introduction of the edit direction $\Delta \boldsymbol{c}$, cross-attention map loss $\mathcal{L}_{c}$, and CUT loss $\mathcal{L}_{e}$, we conduct ablation experiments for comparison. The specific experimental results are presented in Fig.~\ref{tu5}. Our ablation study includes four cases: (1) introducing only the edit direction $\Delta \boldsymbol{c}$, (2) introducing both the edit direction $\Delta \boldsymbol{c}$ and cross-attention map loss $\mathcal{L}_{c}$, (3) introducing the edit direction $\Delta \boldsymbol{c}$, cross-attention map loss $\mathcal{L}_{c}$, and CUT loss $\mathcal{L}_{e}$, and (4) sampling only with DDIM using word swap. Based on the experimental results, we draw the following conclusions: First, introducing the edit direction $\Delta \boldsymbol{c}$ yields a clearer and more realistic target in the generated image. Second, the inclusion of the cross-attention map loss $\mathcal{L}_{c}$ ensures that the target position in the generated image aligns with the target position in the reference image. Third, incorporating the CUT loss $\mathcal{L}_{e}$ helps retain more of the content structure from the reference image in the generated output.

\begin{figure} 
	\centering
	\includegraphics[width=\columnwidth]{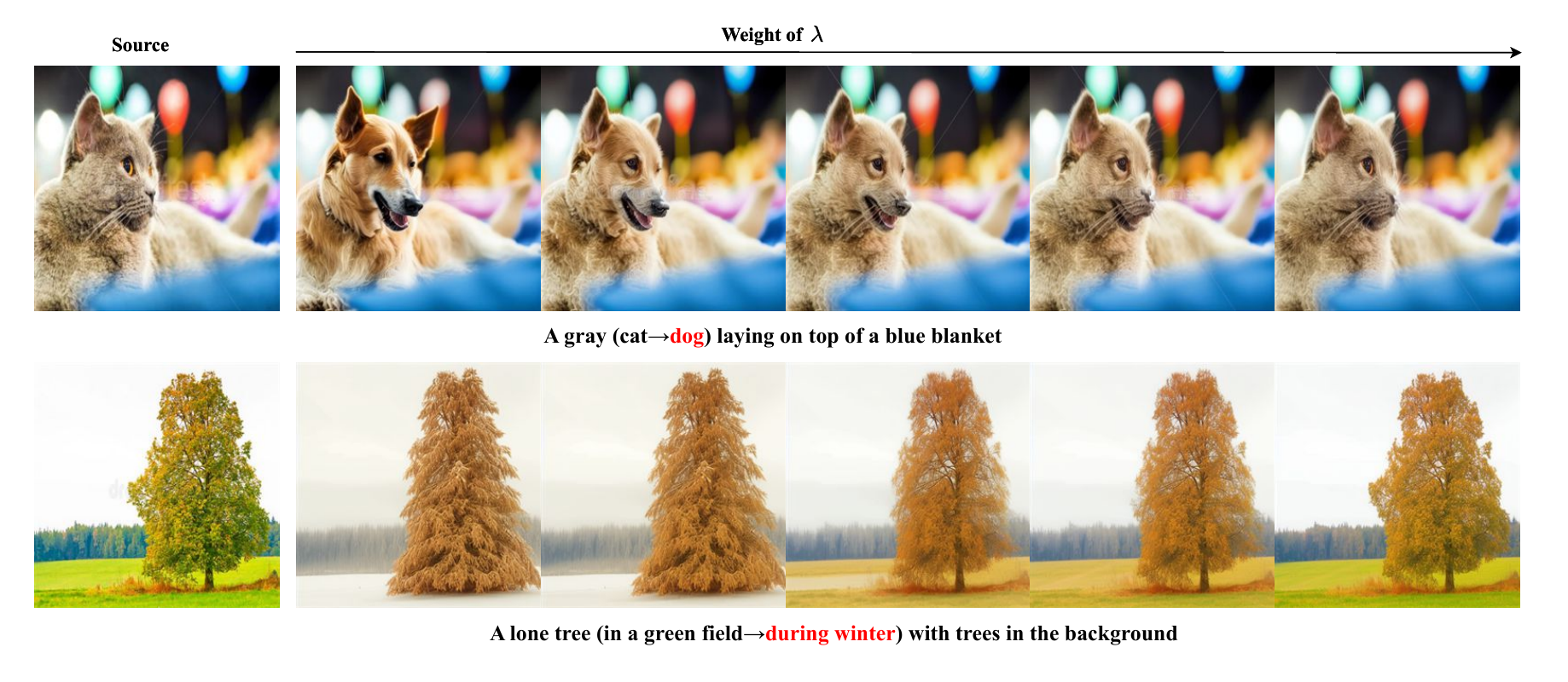}
	\caption{The effect of different learning rate $\lambda$ on our model editing results.}
	\label{tuw}
\end{figure}

\noindent \textbf{Impact of loss weight.}To further investigate the impact of our model's weight parameters on editing outcomes, we conducted additional experiments. Based on our empirical observations, the learning rate parameter $\lambda$ exerts a significant influence on the model’s editing performance; therefore, we focus our analysis on this particular weight. In Fig.~\ref{tuw}, the arrows from left to right represent an increasing value of this weight. The results indicate that when the weight value is smaller, our model not only preserves more structural information from the source image but also produces outputs that better align with the target prompt.

\subsection{Limitations and Discussion}
\begin{figure} 
	\centering
	\includegraphics[width=\columnwidth]{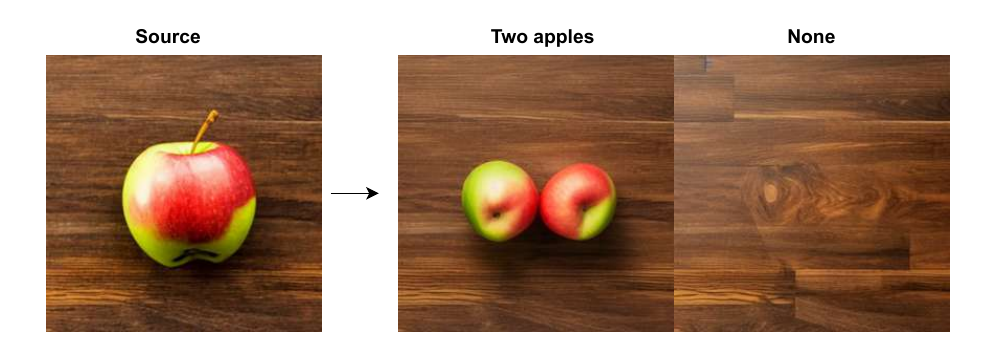}
	\caption{Unsatisfactory results of our method on the target addition and target removal tasks.}
	\label{xianzhi}
\end{figure}

\noindent \textbf{Limitations.} In order to study the editing performance of our method deeply, we here further analyze the limitations of our method by taking the two tasks of object adding and object removing as examples. It should be noted that the source and target prompts are obtained from the pre-trained BLIP model as follows: A apples on the table, Two apples on the table, and None on the table. From the results in Fig.\ref{xianzhi}, it can be seen that although our method can comply with the editing requirements of the target prompts, the generated images are not exactly consistent with the reference image background. In addition, it can be seen that the viewpoints of the two apples in the generated image are also inconsistent with the viewpoints of the apples in the reference image, which further motivates our interest in further investigating ways to solve this problem.

\section{Conclusion}
In this paper, we propose a novel diffusion model-based approach for image-to-image translation tasks by incorporating the contrastive unpaired translation (CUT) loss. The key advantages of our approach are summarized in three points: First, it eliminates the need for users to manually provide target prompts, significantly reducing user effort. Second, our method requires no additional training, leading to substantial savings in computational overhead. Third, by introducing the CUT loss, our method enhances the model’s ability to preserve the content structure of the reference image in the generated output. Additionally, we demonstrate the performance of the proposed model across various image-to-image translation tasks. A limitation of our approach is its inability to fully preserve the content structure of the reference image in tasks involving target removal or addition, which will be a focus of our future research.

\section*{Acknowledgments}
This work is supported by the National Natural Science Foundation of China (62472137, 62072151), Anhui Provincial Natural Science Fund for the Distinguished Young Scholars (2008085J30), Open Foundation of Yunnan Key Laboratory of Software Engineering (2023SE103),  CCF-Baidu Open Fund (CCF-BAIDU202321) and CAAI-Huawei MindSpore Open Fund (CAAIXSJLJJ-2022-057A). Zhao Zhang is the corresponding author of this paper. 

\bibliography{reference}

\begin{thebibliography}{10}
\providecommand{\url}[1]{#1}
\csname url@samestyle\endcsname
\providecommand{\newblock}{\relax}
\providecommand{\bibinfo}[2]{#2}
\providecommand{\BIBentrySTDinterwordspacing}{\spaceskip=0pt\relax}
\providecommand{\BIBentryALTinterwordstretchfactor}{4}
\providecommand{\BIBentryALTinterwordspacing}{\spaceskip=\fontdimen2\font plus
\BIBentryALTinterwordstretchfactor\fontdimen3\font minus
  \fontdimen4\font\relax}
\providecommand{\BIBforeignlanguage}[2]{{%
\expandafter\ifx\csname l@#1\endcsname\relax
\typeout{** WARNING: IEEEtran.bst: No hyphenation pattern has been}%
\typeout{** loaded for the language `#1'. Using the pattern for}%
\typeout{** the default language instead.}%
\else
\language=\csname l@#1\endcsname
\fi
#2}}
\providecommand{\BIBdecl}{\relax}
\BIBdecl

\bibitem{LDM1}
A.~Ramesh, P.~Dhariwal, A.~Nichol, C.~Chu, and M.~Chen, ``Hierarchical
  text-conditional image generation with clip latents,'' \emph{arXiv preprint
  arXiv:2204.06125}, vol.~1, no.~2, p.~3, 2022.

\bibitem{LDM2}
R.~Rombach, A.~Blattmann, D.~Lorenz, P.~Esser, and B.~Ommer, ``High-resolution
  image synthesis with latent diffusion models,'' in \emph{Proceedings of the
  IEEE/CVF conference on computer vision and pattern recognition}, 2022, pp.
  10\,684--10\,695.

\bibitem{sd}
------, ``High-resolution image synthesis with latent diffusion models,'' in
  \emph{Proceedings of the IEEE/CVF conference on computer vision and pattern
  recognition}, 2022, pp. 10\,684--10\,695.

\bibitem{sd1}
A.~Nichol, P.~Dhariwal, A.~Ramesh, P.~Shyam, P.~Mishkin, B.~McGrew,
  I.~Sutskever, and M.~Chen, ``Glide: Towards photorealistic image generation
  and editing with text-guided diffusion models,'' \emph{arXiv preprint
  arXiv:2112.10741}, 2021.

\bibitem{sd2}
J.~Yu, Y.~Xu, J.~Y. Koh, T.~Luong, G.~Baid, Z.~Wang, V.~Vasudevan, A.~Ku,
  Y.~Yang, B.~K. Ayan \emph{et~al.}, ``Scaling autoregressive models for
  content-rich text-to-image generation,'' \emph{arXiv preprint
  arXiv:2206.10789}, vol.~2, no.~3, p.~5, 2022.

\bibitem{sd3}
A.~Ramesh, P.~Dhariwal, A.~Nichol, C.~Chu, and M.~Chen, ``Hierarchical
  text-conditional image generation with clip latents,'' \emph{arXiv preprint
  arXiv:2204.06125}, vol.~1, no.~2, p.~3, 2022.

\bibitem{2}
T.~Brooks, A.~Holynski, and A.~A. Efros, ``Instructpix2pix: Learning to follow
  image editing instructions,'' in \emph{Proceedings of the IEEE/CVF conference
  on computer vision and pattern recognition}, 2023, pp. 18\,392--18\,402.

\bibitem{pnp}
N.~Tumanyan, M.~Geyer, S.~Bagon, and T.~Dekel, ``Plug-and-play diffusion
  features for text-driven image-to-image translation,'' in \emph{Proceedings
  of the IEEE/CVF Conference on Computer Vision and Pattern Recognition}, 2023,
  pp. 1921--1930.

\bibitem{P2P}
A.~Hertz, R.~Mokady, J.~Tenenbaum, K.~Aberman, Y.~Pritch, and D.~Cohen-Or,
  ``Prompt-to-prompt image editing with cross attention control,'' \emph{arXiv
  preprint arXiv:2208.01626}, 2022.

\bibitem{p2pzero}
G.~Parmar, K.~Kumar~Singh, R.~Zhang, Y.~Li, J.~Lu, and J.-Y. Zhu, ``Zero-shot
  image-to-image translation,'' in \emph{ACM SIGGRAPH 2023 Conference
  Proceedings}, 2023, pp. 1--11.

\bibitem{19}
T.~Park, A.~A. Efros, R.~Zhang, and J.-Y. Zhu, ``Contrastive learning for
  unpaired image-to-image translation,'' in \emph{Computer Vision--ECCV 2020:
  16th European Conference, Glasgow, UK, August 23--28, 2020, Proceedings, Part
  IX 16}.\hskip 1em plus 0.5em minus 0.4em\relax Springer, 2020, pp. 319--345.

\bibitem{28}
S.~Yang, H.~Hwang, and J.~C. Ye, ``Zero-shot contrastive loss for text-guided
  diffusion image style transfer,'' in \emph{Proceedings of the IEEE/CVF
  International Conference on Computer Vision}, 2023, pp. 22\,873--22\,882.

\bibitem{DiffuseIT}
\BIBentryALTinterwordspacing
G.~Kwon and J.~C. Ye, ``Diffusion-based image translation using disentangled
  style and content representation,'' 2023. [Online]. Available:
  \url{https://arxiv.org/abs/2209.15264}
\BIBentrySTDinterwordspacing

\bibitem{DDS}
A.~Hertz, K.~Aberman, and D.~Cohen-Or, ``Delta denoising score,'' in
  \emph{Proceedings of the IEEE/CVF International Conference on Computer
  Vision}, 2023, pp. 2328--2337.

\bibitem{CDS}
H.~Nam, G.~Kwon, G.~Y. Park, and J.~C. Ye, ``Contrastive denoising score for
  text-guided latent diffusion image editing,'' in \emph{Proceedings of the
  IEEE/CVF conference on computer vision and pattern recognition}, 2024, pp.
  9192--9201.

\bibitem{BLIP}
D.~Li, J.~Li, and S.~Hoi, ``Blip-diffusion: Pre-trained subject representation
  for controllable text-to-image generation and editing,'' \emph{Advances in
  Neural Information Processing Systems}, vol.~36, pp. 30\,146--30\,166, 2023.

\bibitem{CLIP}
A.~Radford, J.~W. Kim, C.~Hallacy, A.~Ramesh, G.~Goh, S.~Agarwal, G.~Sastry,
  A.~Askell, P.~Mishkin, J.~Clark \emph{et~al.}, ``Learning transferable visual
  models from natural language supervision,'' in \emph{International conference
  on machine learning}.\hskip 1em plus 0.5em minus 0.4em\relax PmLR, 2021, pp.
  8748--8763.

\bibitem{7}
T.~Chen, M.-M. Cheng, P.~Tan, A.~Shamir, and S.-M. Hu, ``Sketch2photo: Internet
  image montage,'' \emph{ACM transactions on graphics (TOG)}, vol.~28, no.~5,
  pp. 1--10, 2009.

\bibitem{10}
T.~Dekel, C.~Gan, D.~Krishnan, C.~Liu, and W.~T. Freeman, ``Sparse, smart
  contours to represent and edit images,'' in \emph{Proceedings of the IEEE
  conference on computer vision and pattern recognition}, 2018, pp. 3511--3520.

\bibitem{17}
C.~Jacobs, D.~Salesin, N.~Oliver, A.~Hertzmann, and A.~Curless, ``Image
  analogies,'' in \emph{Proceedings of SIGGRAPH}, 2001, pp. 327--340.

\bibitem{23}
K.~Kim, S.~Park, E.~Jeon, T.~Kim, and D.~Kim, ``A style-aware discriminator for
  controllable image translation,'' in \emph{Proceedings of the IEEE/CVF
  conference on computer vision and pattern recognition}, 2022, pp.
  18\,239--18\,248.

\bibitem{29}
T.~Park, A.~A. Efros, R.~Zhang, and J.-Y. Zhu, ``Contrastive learning for
  unpaired image-to-image translation,'' in \emph{Computer Vision--ECCV 2020:
  16th European Conference, Glasgow, UK, August 23--28, 2020, Proceedings, Part
  IX 16}.\hskip 1em plus 0.5em minus 0.4em\relax Springer, 2020, pp. 319--345.

\bibitem{30}
T.~Park, J.-Y. Zhu, O.~Wang, J.~Lu, E.~Shechtman, A.~Efros, and R.~Zhang,
  ``Swapping autoencoder for deep image manipulation,'' \emph{Advances in
  Neural Information Processing Systems}, vol.~33, pp. 7198--7211, 2020.

\bibitem{1}
Y.~Alaluf, O.~Patashnik, and D.~Cohen-Or, ``Restyle: A residual-based stylegan
  encoder via iterative refinement,'' in \emph{Proceedings of the IEEE/CVF
  international conference on computer vision}, 2021, pp. 6711--6720.

\bibitem{45}
O.~Tov, Y.~Alaluf, Y.~Nitzan, O.~Patashnik, and D.~Cohen-Or, ``Designing an
  encoder for stylegan image manipulation,'' \emph{ACM Transactions on Graphics
  (TOG)}, vol.~40, no.~4, pp. 1--14, 2021.

\bibitem{35}
E.~Richardson, Y.~Alaluf, O.~Patashnik, Y.~Nitzan, Y.~Azar, S.~Shapiro, and
  D.~Cohen-Or, ``Encoding in style: a stylegan encoder for image-to-image
  translation,'' in \emph{Proceedings of the IEEE/CVF conference on computer
  vision and pattern recognition}, 2021, pp. 2287--2296.

\bibitem{46}
N.~Tumanyan, O.~Bar-Tal, S.~Bagon, and T.~Dekel, ``Splicing vit features for
  semantic appearance transfer,'' in \emph{Proceedings of the IEEE/CVF
  Conference on Computer Vision and Pattern Recognition}, 2022, pp.
  10\,748--10\,757.

\bibitem{39}
C.~Saharia, W.~Chan, H.~Chang, C.~Lee, J.~Ho, T.~Salimans, D.~Fleet, and
  M.~Norouzi, ``Palette: Image-to-image diffusion models,'' in \emph{ACM
  SIGGRAPH 2022 conference proceedings}, 2022, pp. 1--10.

\bibitem{48}
T.~Wang, T.~Zhang, B.~Zhang, H.~Ouyang, D.~Chen, Q.~Chen, and F.~Wen,
  ``Pretraining is all you need for image-to-image translation,'' \emph{arXiv
  preprint arXiv:2205.12952}, 2022.

\bibitem{SG}
J.-Y. Zhu, T.~Park, P.~Isola, and A.~A. Efros, ``Unpaired image-to-image
  translation using cycle-consistent adversarial networks,'' in
  \emph{Proceedings of the IEEE international conference on computer vision},
  2017, pp. 2223--2232.

\bibitem{CDS3}
X.~Huang, M.-Y. Liu, S.~Belongie, and J.~Kautz, ``Multimodal unsupervised
  image-to-image translation,'' in \emph{Proceedings of the European conference
  on computer vision (ECCV)}, 2018, pp. 172--189.

\bibitem{CDS2}
H.~Fu, M.~Gong, C.~Wang, K.~Batmanghelich, K.~Zhang, and D.~Tao,
  ``Geometry-consistent generative adversarial networks for one-sided
  unsupervised domain mapping,'' in \emph{Proceedings of the IEEE/CVF
  conference on computer vision and pattern recognition}, 2019, pp. 2427--2436.

\bibitem{CDS1}
S.~Benaim and L.~Wolf, ``One-sided unsupervised domain mapping,''
  \emph{Advances in neural information processing systems}, vol.~30, 2017.

\bibitem{14}
G.~Kwon and J.~C. Ye, ``One-shot adaptation of gan in just one clip,''
  \emph{IEEE Transactions on Pattern Analysis and Machine Intelligence},
  vol.~45, no.~10, pp. 12\,179--12\,191, 2023.

\bibitem{DDPM}
J.~Ho, A.~Jain, and P.~Abbeel, ``Denoising diffusion probabilistic models,''
  \emph{Advances in neural information processing systems}, vol.~33, pp.
  6840--6851, 2020.

\bibitem{DDIM}
J.~Song, C.~Meng, and S.~Ermon, ``Denoising diffusion implicit models,''
  \emph{arXiv preprint arXiv:2010.02502}, 2020.

\bibitem{GPT3}
T.~Brown, B.~Mann, N.~Ryder, M.~Subbiah, J.~D. Kaplan, P.~Dhariwal,
  A.~Neelakantan, P.~Shyam, G.~Sastry, A.~Askell \emph{et~al.}, ``Language
  models are few-shot learners,'' \emph{Advances in neural information
  processing systems}, vol.~33, pp. 1877--1901, 2020.

\bibitem{masatrol}
M.~Cao, X.~Wang, Z.~Qi, Y.~Shan, X.~Qie, and Y.~Zheng, ``Masactrl: Tuning-free
  mutual self-attention control for consistent image synthesis and editing,''
  in \emph{Proceedings of the IEEE/CVF International Conference on Computer
  Vision}, 2023, pp. 22\,560--22\,570.

\bibitem{freecustom}
G.~Ding, C.~Zhao, W.~Wang, Z.~Yang, Z.~Liu, H.~Chen, and C.~Shen, ``Freecustom:
  Tuning-free customized image generation for multi-concept composition,'' in
  \emph{Proceedings of the IEEE/CVF Conference on Computer Vision and Pattern
  Recognition}, 2024, pp. 9089--9098.

\bibitem{Image-R}
D.~Hendrycks, S.~Basart, N.~Mu, S.~Kadavath, F.~Wang, E.~Dorundo, R.~Desai,
  T.~Zhu, S.~Parajuli, M.~Guo \emph{et~al.}, ``The many faces of robustness: A
  critical analysis of out-of-distribution generalization,'' in
  \emph{Proceedings of the IEEE/CVF international conference on computer
  vision}, 2021, pp. 8340--8349.

\bibitem{LAION-5B}
C.~Schuhmann, R.~Beaumont, R.~Vencu, C.~Gordon, R.~Wightman, M.~Cherti,
  T.~Coombes, A.~Katta, C.~Mullis, M.~Wortsman \emph{et~al.}, ``Laion-5b: An
  open large-scale dataset for training next generation image-text models,''
  \emph{Advances in Neural Information Processing Systems}, vol.~35, pp.
  25\,278--25\,294, 2022.

\bibitem{BGLIPIS}
X.~Zhou, R.~Girdhar, A.~Joulin, P.~Kr{\"a}henb{\"u}hl, and I.~Misra,
  ``Detecting twenty-thousand classes using image-level supervision,'' in
  \emph{European Conference on Computer Vision}.\hskip 1em plus 0.5em minus
  0.4em\relax Springer, 2022, pp. 350--368.

\bibitem{Dist}
M.~Caron, H.~Touvron, I.~Misra, H.~J{\'e}gou, J.~Mairal, P.~Bojanowski, and
  A.~Joulin, ``Emerging properties in self-supervised vision transformers,'' in
  \emph{Proceedings of the IEEE/CVF international conference on computer
  vision}, 2021, pp. 9650--9660.

\end{thebibliography}
\bibliographystyle{IEEEtran}

\end{document}